# Causal Networks: Semantics and Expressiveness*

Thomas Verma & Judea Pearl
Cognitive Systems Laboratory, Computer Science Department
University of California Los Angeles, CA 90024
<verma@cs.ucla.edu> <judea@cs.ucla.edu>

## ABSTRACT

Dependency knowledge of the form "$x$ is independent of $y$ once $z$ is known" invariably obeys a set of four axioms defining *semi-graphoids*, examples of which are probabilistic and database dependencies. Such knowledge can often be stored efficiently in graphical structures, using either undirected graphs and directed acyclic graphs (DAGs). This paper shows that, if we construct a DAG from any causal input list of a semi-graphoid, then a graphical criterion called *d-separation* is a sound rule for reading independencies from the DAG, i.e., it produces only valid assertions of conditional independence. The rule is extended to include DAGs with functional dependencies.

## INTRODUCTION

Dependency knowledge is useful in several areas of research, for example in database design it is useful to reason about embedded-multivalued-dependence (EMVD) of attributes [Fagin, 1977] and in decision analysis and expert systems design it is useful to reason about probabilistic independence of variables [Howard and Matheson, 1981], [Pearl, 1988]. These represent two formalizations of the intuitive relation "knowing Z renders X and Y independent" which shall be denoted as $I(X, Z, Y)$. This relation would naturally have different properties in different applications, but it is interesting to note that most sensible definitions of this relation share four common properties listed below:

| | | |
|---|---|---|
| symmetry | $I(X, Z, Y) \Leftrightarrow I(Y, Z, X)$ | (1.a) |
| decomposition | $I(X, Z, YW) \Rightarrow I(X, Z, Y)$ | (1.b) |
| weak union | $I(X, Z, YW) \Rightarrow I(X, ZY, W)$ | (1.c) |
| contraction | $I(X, ZY, W) \& I(X, Z, Y) \Rightarrow I(X, Z, YW)$ | (1.d) |

where $X$, $Y$ and $Z$ represent three disjoint subsets of objects (e.g. variables, attributes). It is known that every EMVD relation obeys the four properties listed above, as well as many other properties. (The notation $I(X, Z, Y)$ is equivalent to the standard EMVD notation $Z \twoheadrightarrow X \mid Y$). Probabilistic dependencies also obey these four properties [Dawid, 1979], and it has been conjectured that they are, in fact, complete [Pearl and Paz, 1985], namely, that any other property of probabilistic independence is a logical consequence of the four. Three place relations which obey these four properties are called *semi-graphoids*.

---

* This work was partially supported by the National Science Foundation Grant #IRI-8610155. "Graphoids: A Computer Representation for Dependencies and Relevance in Automated Reasoning (Computer Information Science)."

For probability distributions containing strictly positive probabilities, the independence relation has a fifth independent property:

$$\text{intersection} \quad I(X, ZY, W) \ \& \ I(X, ZW, Y) \Rightarrow I(X, Z, YW) \quad (2)$$

This property along with the four of semi-graphoids define the class of *graphoids* (also conjectured to be complete for non-extreme probabilities [Pearl and Paz, 1985]).

A naive approach for representing a dependency model, i.e., particular instance of a dependency relation, would be to enumerate all triplets $(X, Z, Y)$ for which $I(X, Z, Y)$ holds. This could result in exponential space since the relation $I$ ranges over subsets of objects. The use of graphs as a representation of dependency models is appealing in three ways; first the graph has an intuitive conceptual meaning, second, it is an efficient representation in terms of time and space [Pearl and Verma, 1987], third there are efficient methods which utilize the information when it is organized this way [Pearl, 1982, 1985, 1986, 1988], [Shachter, 1985, 1988].

## UNDIRECTED GRAPHS

The meaning of a particular undirected graph is straight forward, each node in the graph represents a variable, and a link in the graph means that the two variables are directly dependent. Under this meaning, a set of nodes $Z$ would *separate* two other sets $X$ and $Y$, if and only if every path between a node in $X$ and a node in $Y$ passes through $Z$. This representation can fully represent only a small set of dependency models defined by the following properties [Pearl and Paz, 1985]:

| | | |
|---|---|---|
| symmetry | $I(X, Z, Y) \Leftrightarrow I(Y, Z, X)$ | (3.a) |
| decomposition | $I(X, Z, YW) \Rightarrow I(X, Z, Y)$ | (3.b) |
| strong union | $I(X, Z, Y) \Rightarrow I(X, ZW, Y)$ | (3.c) |
| intersection | $I(X, ZY, W) \ \& \ I(X, ZW, Y) \Rightarrow I(X, Z, YW)$ | (3.d) |
| transitivity | $I(X, Z, Y) \Rightarrow I(X, Z, \gamma) \text{ or } I(Y, Z, \gamma) \ \forall \gamma \notin X \cup Y \cup Z$ | (3.e) |

It is not always necessary to have an exact graphical representation of a dependency model, in fact an efficient approximation called an *I-map* is often preferred to an inefficient perfect map. A representation $R$ of a dependency model $M$ is an I-map if every independence represented in $R$ implies a valid independence in $M$. Thus, $R$ may not contain all independencies of $M$, but the ones it does contain are correct. There is an algorithm which finds the most representative I-map for any graphoid [Pearl and Paz, 1985]. Since probabilistic independence over positive probabilities constitutes a graphoid, there is always a unique *edge-minimal* undirected graph which is an I-map of any non-extreme probabilistic distribution $P$ (i.e., no edge can be removed without destroying the I-mapness of the graph). This is not the case for EMVD relations nor for probabilistic distributions that exclude certain combinations of events; in these cases there is no unique edge-minimal I-map for a given model, and, moreover, there is no effective method of constructing even one minimal I-map.

## DIRECTED-ACYCLIC GRAPHS (DAGS)

The dependency model represented by a particular DAG has a simple causal interpretation; each node represents a variable and there is a directed arc from one node to another if the first is a direct cause of the second. Under this interpretation, graph-separation is not as straight forward as before since two unrelated causes of a symptom may become related once the symptom is observed [Pearl, 1986]. Thus a set of nodes $Z$ is defined to *d-separate* two other sets $X$ and $Y$ if and only if every *adjacency path* from a node in $X$ to a node in $Y$ is rendered *inactive* by $Z$. An adjacency path is a path which follows arcs ignoring their directionality; one is rendered *inactive* by a set of nodes $Z$ if and only if either there is a *head-to-head* node along the path which is not in $Z$ and none of its descendents are in $Z$ or some node along the path is not head-to-head but is in $Z$. A node along the path is head-to-head if the node before it and after it along the path both point to it in the graph. One node is a descendent of another if there is a directed path from the latter to the former.

There is a procedure which produces a minimal I-map of any dependency model which is a semi-graphoid. It employs an algorithm which takes a *stratified protocol* (also called *causal input list* [Geiger and Pearl, 1988]) of a dependency model and produces a perfect map of its semi-graphoid closure. A stratified protocol of a dependency model contains two things: an ordering of the variables, and a function that assigns a *tail boundary* to each variable $x$. A tail boundary of a variable $x$ is any set of lesser variables (with respect to the ordering) rendering $x$ independent of all other lesser variables. A unique DAG can be generated from each stratified protocol by associating the set of direct parents of any node $x$ in the DAG with the tail boundary of the variable $x$ in the protocol. An equivalent specification of a stratified protocol is an ordered list of triplets of the form $I(n, B, R)$, one triplet for each variable in the model, where the set $B$ is the tail boundary of the variable $n$ and $R$ is a set containing all other lesser variables.

For a particular dependency model over $n$ variables there are $n!$ orderings, and for each ordering there can be up to $\prod_{k=1}^{n} 2^{k-1} = 2^{n(n-1)/2}$ different sets of tail boundaries since, in the worst case, every subset of lesser variables could be a boundary. Thus, there can be as many as $n! \, 2^{n(n-1)/2}$ stratified protocols. But if the dependency model posses a perfect map in DAGs, then one of the protocols is guaranteed to generate it by the following theorem.

**Theorem 1:** If $M$ is a dependency model which can be perfectly represented by some DAG $D$, then there is a stratified protocol $L_\theta$ which generates D.

**Proof:** Let $D$ be a DAG which perfectly represents $M$. Since $D$ is a directed acyclic graph it imposes a partial order $\phi$ on the variables of $M$. Let $\theta$ be any total ordering consistent with $\phi$ (i.e. $a <_\phi b \Rightarrow a <_\theta b$). For any node $n$ in $D$, the set of its parents $P(n)$ constitutes a tail boundary with respect to the ordering $\theta$, thus the pair $L_\theta = (\theta, P(n))$ is a stratified protocol of $M$, and this is the very protocol which will generate $D$. QED.

Although it is possible to find a perfect map when it exists, testing for existence may be intractable. It is practical, however, to find a minimal I-map, and the next theorem shows that stratified protocols can be used to generate I-maps of any semi-graphoid, not necessarily those possessing perfect maps in DAGs.

354

**Theorem 2:** If $M$ is a semi-graphoid, and $L_\theta$ is any stratified protocol of $M$, then the DAG generated by $L_\theta$ is an I-map of $M$.

**Proof:** Induct on the number of variables in the semi-graphoid. For semi-graphoids of one variable it is obvious that the DAG generated is an I-map. Suppose for semi-graphoids with fewer than $k$ variables that the DAG is also an I-map. Let $M$ have $k$ variables, $n$ be the last variable in the ordering $\theta$, $M-n$ be the semi-graphoid formed by removing $n$ and all triplets involving $n$ from $M$ and $G-n$ be the DAG formed by removing $n$ and all its incident links from $G$. Since $n$ is the last variable in the ordering, it cannot appear in any of boundaries of $L_\theta$, and thus $L_\theta - n$ can be defined to contain only the first $n-1$ variables and boundaries of $L_\theta$ and still be a stratified protocol of $M-n$. In fact the DAG generated from $L_\theta - n$ is $G-n$. Since $M-n$ has $k-1$ variables, $G-n$ is an I-map of it. Let $M_G$ be the dependency model corresponding to the DAG $G$, and $M_{G-n}$ correspond to $G-n$, (i.e. $M_G$ contains all d-separated triplets of $G$).

$G$ is an I-map of $M$ if and only if $M_G \subseteq M$. Each triplet $T$ of $M_G$ falls into one of four categories; either the variable $n$ does not appear in $T$ or it appears in the first, second or third entry of $T$. These will be treated separately as cases 1, 2, 3 and 4, respectively.

**case-1:** If $n$ does not appear in $T$ then $T$ must equal $(X, Z, Y)$ with $X, Y$ and $Z$ three disjoint subsets of variables, none of which contain $n$. Since $T$ is in $M_G$ it must also be in $M_{G-n}$ for if it were not then there would be an active path in $G-n$ between a node in $X$ and a node in $Y$ when $Z$ is instantiated. But if this path is active in $G-n$ then it must also be active in $G$ since the addition of nodes and links can not deactivate a path. Since $G-n$ is an I-map of $M-n$, $T$ must also be an element of it, but $M-n$ is a subset of $M$, so $T$ is in $M$

**case-2:** If $n$ appears in the first entry of the triplet, then $T = (Xn, Z, Y)$ with the same constraints on $X, Y$ and $Z$ as in case-1. Let $(n, B, R)$ be the last triple in $L_\theta$, $B_X, B_Y, B_Z$ and $B_0$ be a partitioning of $B$ and $R_X$, $R_Y, R_Z$ and $R_0$ be a partitioning of $R$ such that $X = B_X \cup R_X$, $Y = B_Y \cup R_Y$ and $Z = B_Z \cup R_Z$ as in figure 1.

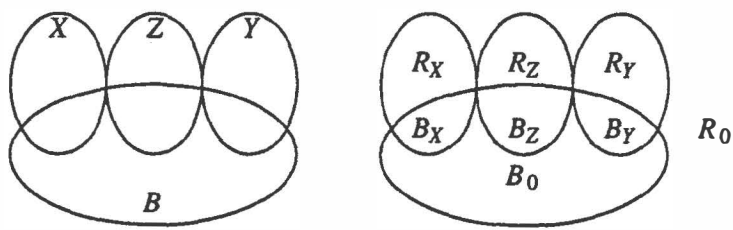

Figure 1

By the method of construction, there is an arrow from every node in $B$ to $n$, but since $(Xn, Z, Y)$ is in $M_G$ every path from a node in $Y$ to $n$ must be deactivated by $Z$ so $B_Y$ must be empty or else there would be a direct link from $Y$ to $n$ (see figure 2a). The last triplet in $L_\theta$ can now be written as $(n, B_X B_0 B_Z, R_X R_Z Y R_0)$. Since $X = B_X \cup R_X$, $Y = R_Y$ and $M$ is a semi-graphoid it follows (from (1.b) and (1.c)) that $(n, XB_0 Z, Y) \in M$.

355

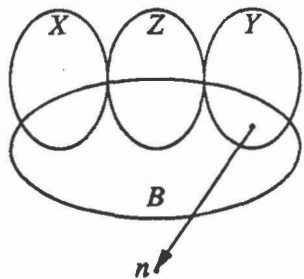 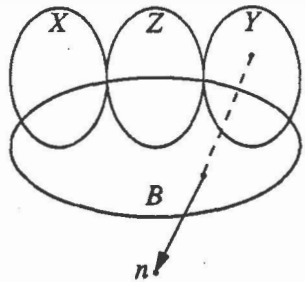

Figure 2a                Figure 2b

Since there is an arrow from every node in $B_0$ to $n$ and $n$ is separated from $Y$ given $Z$ in $G$, $B_0$ must also be d-separated from $Y$ given $Z$ in $G$ for if it were connected there would be a path from a node in $Y$ to a node in $B_0$ which was active given $Z$. But there is an arrow from every node in $B_0$ to $n$, thus, such a path would also connect the node in $Y$ to $n$, and $Y$ would no longer be separated from $n$ given $Z$ (see figure 2b). Since $Y$ is separated from both $B_0$ and $X$ given $Z$ in the DAG $G$ it is separated from their union, so $(XB_0, Z, Y) \in M_G$. Since $n$ is not in this triplet, the argument of case-1 above implies that $(XB_0, Z, Y) \in M$. Since $(n, XB_0Z, Y) \in M$ and $M$ is a semi-graphoid it follows (using (1.b) and (1.d)) that $T = (Xn, Z, Y) \in M$

**case-3:** If $n$ appears in the second entry then $T = (X, Zn, Y)$ with the same constraints on $X$, $Y$ and $Z$ as in case-1. Also let $B$, $R$, etc. be defined as in case-2, thus $(n, B_X B_Y B_Z B_0, R_X R_Y R_Z R_0) \in M$ since it is the last triplet in $L_\theta$.

In this case, either $B_y$ is empty and $B_0$ is separated from $Y$ given $Z$ in $G$, or $B_X$ is empty and $B_0$ is separated from $X$ given $Z$ in $G$ since if neither were the case, then there would be a path from a node in $Y$ which would be active given $Z$ and would end *pointing* at $n$, and there would be a similar path from a node in $X$ to $n$. But this means that there would be a path from a node in $X$ to a node in $Y$ which would be active given $Z$ and $n$ since the path is head-to-head at $n$ (see figure 3a and 3b). But there can be no such path since by assumption $(X, Zn, Y) \in M_G$. Without loss of generality, assume that $B_Y = \emptyset$ and $(B_0, Z, Y) \in M_G$.

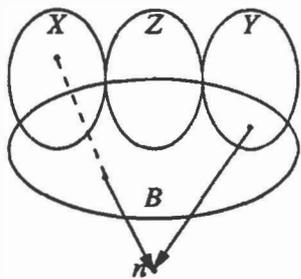 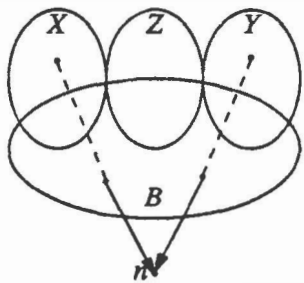

Figure 3a                Figure 3b

$X$ and $Y$ must be separated in $G$ given only $Z$ for if they were not then there would be a path between them

which would be active given $Z$. Since they are separated given $Z$ and $n$, this path would have to be deactivated by $n$, but since there are only arrows pointing at $n$ it can only activate paths by being instantiated (see figure 3b). Thus, there can be no such path and $X$ and $Y$ must be separated given $Z$ in $G$. Since $Y$ is separated from both $X$ and $B_0$ given $Z$ in $G$ it follows that $(XB_0, Z, Y) \in M_G$ and by the argument of case-1 above $(XB_0, Z, Y) \in M$. Since $B_Y$ is empty it follows, as in case-2, that $(n, XB_0 Z, Y) \in M$. Further, $M$ is a semi-graphoid so, applying (1.d) to $(XB_0, Z, Y) \in M$ and $(n, XB_0 Z, Y) \in M$ yields $(nXB_0, Z, Y) \in M$ and using (1.b) and (1.c) it follows that $T = (X, Zn, Y) \in M$.

**case-4:** If $n$ appears in the third entry, then by symmetry the triplet $T$ is equivalent to one with $n$ in the first entry, and the argument of case-2 above shows that $T \in M$. QED.

**Corollary 1:** If $L_\theta$ is any stratified protocol of some dependency model $M$, the DAG generated from $L_\theta$ is a perfect map of the semi-graphoid closure of $L_\theta$. In other words, a triplet is d-separated in the DAG if and only if it can be derived from the triplets of $L_\theta$ using the four axioms in (1).

**Proof:** By the previous theorem, the DAG is an I-map of the closure, and it remains to show that the closure is an I-map of the DAG. Since every DAG dependency model is a semi-graphoid, the DAG closure of $L_\theta$ contains the semi-graphoid closure of it, thus, it suffices to show that the DAG dependency model $M_G$ contains $L_\theta$. If $(n, B, R)$ is a triplet in $L_\theta$ then $n$ is separated from $R$ given $B$ in the DAG, for if not then there would be a path from a node in $R$ to $n$ which is active given $B$. But since every link into $n$ is from $B$ the path must lead out of $n$ into some node which was placed after $n$. Since every node in $R$ was placed before $n$, this path cannot be directed and must contain a head-to-head node at some node which was placed after $n$. But this path is deactivated by $B$ since it contains no nodes placed after $n$, and thus, $B$ would separate $n$ from $R$ in the graph. QED.

**Corollary 2:** If each tail boundary in $L_\theta$ is minimal, the resulting DAG is a minimal I-map of $M$.

Theorem 2 and its corollaries together imply that d-separation is sound and complete for the extraction of independence information from DAGs with respect to their stratified protocols. That is, a conclusion can be read from the graph using d-separation if and only if it follows from application of the semi-graphoid axioms to the stratified protocol. In bayesian networks [Pearl, 1986], for example, any independence which can be read from the graph via d-separation is sound with respect to the probability distribution that it represents. Since the axioms of semi-graphoids have not been shown to be complete for the class of probabilistic dependencies, corollary 1 is not enough to ensure that d-separation will identify more independencies than any other sound criterion. The latter has been shown in [Geiger and Pearl, 1988].

The last theorem, which is of a theoretical nature, states that it is possible to force any particular independence of a semi-graphoid to be represented in an I-map.

**Theorem 3:** If $M$ is any semi-graphoid then the set of DAGs generated from all stratified protocols of $M$ is a perfect map of $M$ if the criterion for separation is that $d$-separation must exist in one of the DAGs.

**Proof:** If there is a separation in one of the DAGs then the corresponding independence must hold in $M$ since theorem 2 states that each of the DAGs is an I-map of $M$, thus, the set is also an I-map. It remains to show that $M$ is an I-map of the set of DAGs. Let $T = (X, Z, Y)$ be any triplet in $M$ and $X = \{x_1, \ldots, x_n\}$.

357

The triplets $T^* = \{(x_i, x_1 \cdots x_{i-1}Z, Y) \mid 1 \leq i \leq n\}$ must also be in $M$ since they are implied by $T$ using the weak union axiom of semi-graphoids. Furthermore $T$ is in the semi-graphoid closure of $T^*$ since the triplets imply $T$ by use of the contraction axiom. Thus any protocol containing the triplets $T^*$ would generate a DAG containing $T$. Such a protocol need only have an ordering $\theta$ such that the variables of $Y$ and $Z$ are less than those of $X$ which are less than any other variables and that the variables of $X$ are ordered such that $x_i <_\theta x_j$ if and only if $i < j$. The DAG generated by this protocol is in the set of DAGs and therefore the separation holds in the set. QED.

Since there is an effective algorithm for generating an I-map DAG for any semi-graphoid, DAGs would be a useful means of representing EMVD relations as well as probabilistic independence relations. Furthermore if the particular dependency model is stated as a stratified protocol then it can be perfectly represented by a DAG.

## FUNCTIONAL DEPENDENCIES

The ability to represent functional dependencies would be a powerful extension from the point of view of the designer. These dependencies may easily be represented by the introduction of *deterministic nodes* which would correspond to the deterministic variables [Shachter, 1988]. Graphs which contain deterministic nodes represent more information than d-separation is able to extract; but a simple extension of d-separation, called ID-separation, is both sound and complete with respect to the input protocol under both probabilistic inference and semi-graphoid inference [Geiger and Verma, 1988]. ID-separation is very similar to d-separation, only differing in that a path is rendered *inactive* by a set of nodes $Z$ under ID-separation just in the case that it would be inactive under d-separation plus the case there is a node on the path which is *determined* by $Z$.

## CONCLUSIONS

This paper shows that the criteria of d-separation and ID-separation are sound. They provide a reliable and efficient method for extracting independence information from DAGs. This information may be used explicitly, for example to help guide a complex reasoning system, or implicitly as in bayesian propagation [Pearl, 1986] and the evaluations of Influence Diagrams [Shachter, 1985, 1988]. The criteria also provide a sound theoretical basis for the analysis of the properties of these graphical representations. For example, the validity of graphical manipulations such as arc reversal and node removal [Howard and Matheson, 1981] [Shachter 1985, 1988][Smith, 1987] can now be affirmed on solid theoretical foundations.

## ACKNOWLEDGMENT

We thank Dan Geiger and Azaria Paz for many valuable discussions, and James Smith for checking the proof of Theorem 2.